\documentclass[a4paper]{article}
\usepackage{ISCSLP2024}
\usepackage{ifthen}
\newboolean{blind}
\usepackage{float}
\usepackage{adjustbox}
\usepackage{multirow}
\usepackage{tabularx}
\usepackage{makecell}
\usepackage{textcomp}
\usepackage{xcolor}

\title{Not All Errors Are Equal: Investigation of Speech Recognition Errors in Alzheimer's Disease Detection}

\name{
	\ifthenelse{\boolean{blind}}{Anonymous to ISCSLP}
	{Jiawen Kang, Junan Li, Jinchao Li, Xixin Wu, Helen Meng}
}

\address{
  \ifthenelse{\boolean{blind}}{Anonymous to ISCSLP}{
  The Chinese University of Hong Kong, Hong Kong SAR
  }
}

\email{
    \ifthenelse{\boolean{blind}}{Anonymous to ISCSLP}
    {\{jwkang, jli, jcli, wuxx, hmmeng\}@se.cuhk.edu.hk}
}

\begin{document}

\maketitle
% \newcommand{\JA}[1]{\textcolor{blue}{(LJA: #1)}}
% \newcommand{\JW}[1]{\textcolor{red}{(KJW: #1)}}
% \newcommand{\JC}[1]{\textcolor{red}{(LJC: #1)}}

% The abstract here must exactly match the abstract entered into the paper submission system
% Index terms are defined in `\keywords` in `main.tex`.

\begin{abstract}
\vspace{-0.2cm}

Automatic Speech Recognition (ASR) plays an important role in speech-based automatic detection of Alzheimer's disease (AD).
However, recognition errors could propagate downstream, potentially impacting the detection decisions.
Recent studies have revealed a non-linear relationship between word error rates (WER) and AD detection performance, where ASR transcriptions with notable errors could still yield AD detection accuracy equivalent to that based on manual transcriptions.
This work presents a series of analyses to explore the effect of ASR transcription errors in BERT-based AD detection systems.
Our investigation reveals that not all ASR errors contribute equally to detection performance.
Certain words, such as stopwords, despite constituting a large proportion of errors, are shown to play a limited role in distinguishing AD.
In contrast, the keywords related to diagnosis tasks exhibit significantly greater importance relative to other words.
These findings provide insights into the interplay between ASR errors and the downstream detection model.

\end{abstract}

\vspace{-0.05cm}

\section{Introduction}
\label{sec:intro}
\vspace{-0.05cm}

% intro AD
Alzheimer’s disease (AD) is a neurodegenerative disorder characterized by progressive cognitive impairment, including deterioration in memory, attention, and executive function.
Due to the irreversible progression of AD pathology~\cite{lynch2020world}, early detection and diagnosis play a pivotal role in facilitating timely intervention and management, conventionally relying on in-person clinical assessments~\cite{nasreddine2005montreal, reilly2004frog}.
With recent progress in spoken language technology, speech-based automatic AD detection has emerged as a promising area due to its potential for more cost-effective and scalable AD screening~\cite{martinez2021ten, meng2023integrated}.
\vspace{-0.05cm}

% Automatic AD detection: SOTA methods (PLMs)
Automatic detection of Alzheimer's Disease (AD) from speech has attracted increasing attention in recent years.
Early research attempted to identify AD by utilizing handcrafted acoustic and linguistic features~\cite{alhanai2017spoken, fraser2016linguistic, weiner2019speech, frankenberg2021verbal}.
For example, Winer and Frankenberg et al.~\cite{weiner2019speech, frankenberg2021verbal} demonstrated that parts-of-speech (POS), word categories, and pause features are highly related to AD.
More recently, deep embedding features from pre-trained models have been widely explored for AD detection, leveraging either speech-based models~\cite{koo2020exploiting, balagopalan2021comparingPretrained, syed2021automated} or text-based models~\cite{balagopalan2020bert, yuan2020disfluencies, martinc2021temporal}.
Among these models, pre-trained language models (PLMs) such as BERT~\cite{devlin2019bert} and RoBERTa~\cite{liu2019roberta}, have shown remarkable performance by leveraging their strong ability to capture rich linguistic and semantic patterns.
These PLMs can be used either as feature extractors~\cite{balagopalan2020bert, syed2020automated, li2021comparative} or directly fine-tuned on the AD detection task~\cite{yuan2020disfluencies}.
Moreover, researchers have made further enhancements to PLMs~\cite{li2023leveraging, wang2022exploring, qiao2021alzheimer}, pushing the state-of-the-art detection accuracy to new heights.
\vspace{-0.05cm}

% Importance of ASR -> joint works of ASR and AD -> research gaps
While PLMs have achieved impressive performance in AD detection, the practical applications of these approaches largely rely on Automatic Speech Recognition (ASR) front-ends to transcribe speech into text, as transcription errors could lead to significant bias in PLM embeddings and alter the decisions of downstream AD detection models.
Many works have investigated tailored ASR systems for AD or other disordered speech.
Early works explored incorporating language models~\cite{zhou2016speech} and Maximum A Posteriori (MAP) adaptation~\cite{mirheidari2019dementia} for ASR in AD detection.
Recent efforts have investigated novel approaches such as data augmentation~\cite{ye2021development, geng2022investigation, jin2023personalized}, domain adaptation~\cite{deng2021bayesian} and neural architectural search~\cite{wang2022conformer}.
Nevertheless, due to the difficulty in collecting in-domain data, there is still a substantial accuracy gap between ASR for disordered speech and normal speech.
\vspace{-0.05cm}

% motivation: relationships between ASR and AD detection
% At the same time, many studies have shown a non-linear relationship between ASR error rates and AD detection accuracies \cite{li2021comparative, wang2022conformer, li2022far, li2024useful}, i.e., ASR models with lower Word Error Rate (WER) may not necessarily lead to better AD detection results, and verse versa.
On the other hand, many studies have shown a non-linear relationship between ASR Word Error Rate (WER) and AD detection accuracies.
For instance, ASR transcription with notable errors (WER \raisebox{-0.55ex}{\textasciitilde}30\%) could yield equivalent AD detection accuracy as manual transcription does~\cite{li2021comparative, li2022far}, while ASR models with lower WER might not necessarily lead to better AD detection results~\cite{li2024useful, wang2022conformer}.
% These findings suggest that error distribution, in contrast to overall error rates, might be another critical factor influencing the downstream AD detection task.
Few studies have investigated the in-depth relationship between ASR errors and AD detection. \cite{zhou2016speech} might be the first work revealing that the performances of ASR and AD detection systems are weakly correlated.
Balagopalan et al.~\cite{balagopalan2020impact} demonstrated that deletion error is effective in the AD detection model using a handcrafted feature set.
Recently, Li et al.~\cite{li2024useful} showed that ASR errors could serve as helpful cues for fine-tuned PLMs to identify speech with AD.
These findings suggest that certain ASR error words might be trivial and harmless to the downstream AD detection task, while others could have a strong effect.
\vspace{-0.05cm}

This work aims to explore the impact of ASR errors on Alzheimer's Disease (AD) detection using a BERT-based model. 
We first implement a subject ASR and AD detection systems which achieves the same accuracy when evaluated with ASR and manual transcriptions. 
We then conducted a thorough analysis based on the ASR errors.
It reveals that not all ASR errors are equally detrimental to downstream AD detection.
Certain words, such as stopwords, constitute a large proportion of errors but are less significant concerning the picture description tasks performed in the ADReSS data used in this study.
In contrast, task-related keywords occupy only 9\% of all ASR errors but have proven to play a pivotal role compared to other words.
These findings provide insights into the interplay between ASR errors and the downstream detection model, benefiting the development of ASR-robust AD detection systems.

\section{Approach}
\vspace{-0.2cm}
\subsection{Data}
\vspace{-0.1cm}

The data used in this work comes from the Alzheimer’s Dementia Recognition Through Spontaneous Speech (ADReSS) Challenge 2020 corpus~\cite{luz2020alzheimer}. This challenge selects a sub-task of Pitt Corpus in the DementiaBank database~\cite{becker1994natural}, which requires all the participants to describe the Cookie Theft picture~\cite{kaplan1983boston}. The ADReSS corpus consists of 156 different English speakers' audio samples with corresponding transcripts. Among them, 78 of the speakers are healthy controls (35 male, 43 female) while the rest are with AD (35 male, 43 female). The corpus is divided into a standard train (108 speakers, about 2 hours) and test (48 speakers, about 1 hour) sets with balanced distributions of age, gender, and disease conditions.

 % as shown in Fig.~\ref{fig:cookie}
% \input{figs/cookie}
\vspace{-0.2cm}
\subsection{Automatic Speech Recognition Models}
\label{sec:asr-setting}
\vspace{-0.1cm}
We employed a tailor-made ASR system based on the Time Delay Neural Network (TDNN).
This model was trained with 1000-hour Librispeech corpus and adapted on 59-hour Pitt corpus~\cite{becker1994natural}.
Additionally, data augmentation, speaker adaptation, and a Transformer language model were adopted to enhance the performance on the in-domain speech.
As a result, the employed system has shown state-of-the-art level performance on the ADReSS corpus.
More details can be found in~\cite{ye2021development}.

% \noindent \textbf{Whisper-ft Model}
% Whisper~\cite{radford2023robust} is a large-scale ASR model trained on 680,000 hours of multilingual and multitask speech data.
% We fine-tuned the Whisper base model with LoRA strategy~\cite{hu2021lora} on the ADReSS standard train set.
% Specifically, we fine-tuned the model for 3 epochs with AdamW optimizer with a batch size of $16$.
% We also used a linear decay learning rate scheduler with a peak learning rate of 1e-2 and $10$ warm-up steps.
% As for the LoRA strategy, we set the rank as $32$, alpha as $64$, and dropout rate as $0.05$.
% Note that one can achieve better WER using other settings but we intentionally retain a certain level of ASR errors for analysis.

\vspace{-0.2cm}
\subsection{AD Detection Models}
\vspace{-0.1cm}

We formulate AD detection as a binary classification problem, i.e. classifying participants as healthy controls or individuals with AD.
The development of the detection system follows work ~\cite{li2021comparative}, which has shown superior detection performance on the ADReSS corpus.
This system utilizes the BERT model~\cite{devlin2019bert} to generate embeddings from the automatically transcribed text. 
Specifically, we take the embedding corresponding to the [CLS] token for each speaker throughout the entire transcript. This process yields a feature vector with a dimensionality of 768, which serves as the input representation for the following analysis.
To avoid the over-fitting problem, we use the Principal Component Analysis (PCA) method to reduce the dimension of the embeddings to 108, which is the size of the training set.
The compressed embeddings are then fed into a Support Vector Machine (SVM) with linear kernel and regularization parameter $C=1$ for classification.
Both PCA and SVM models are fitted with manual transcription.
As for evaluation, we use accuracy scores as the main criterion, where precision, recall, and F1 scores are also used for analysis.

% \vspace{-0.2cm}
% \subsection{Evaluation Protocol}
% \vspace{-0.1cm}

% The systems are evaluated by 10-fold cross-validation (CV) on the training data and tested on ADReSS test data. We ran 10 times 10-fold CV and averaged the resulting scores.
% The scores for evaluating classification performance include accuracy scores (ACC), precision (PRE), recall (REC), and F1 scores with respect to the positive class (AD).

\section{AD Detection with ASR Transcriptions}
\label{sec:base}
The non-linear dependency between ASR WER and AD detection accuracy has been observed in multiple studies~\cite{li2021comparative, wang2022conformer, li2022far, li2024useful}.
To look into this observation, we first set up a subject system using the above-mentioned settings.
Table~\ref{tab:basic_acc} shows the classification results of the subject detection system.
Note that the detection model was fitted with manual transcription, and ASR transcriptions were only used in the test stage.
We can see that even though the ASR system has a WER of $33.9\%$, it still achieves an equivalent detection accuracy of $88\%$ compared to manual transcriptions.
% Aligned with other works, this experiment implies a large proportion of errors may play a marginal role in distinguishing AD participants.

% WER	ACC	P	R	F	A
% Manual	0	0.88	0.85	0.92	0.88	0.87
% TDNN	33.9	0.88	0.82	0.96	0.88	0.88
% Whisper	35.4	0.85	0.77	1	0.87	0.85

% \vspace{-10pt}
\begin{table}[htbp]
\vspace{-5pt}
\centering
\caption{AD detection performance using Manual and ASR transcription.
``Trans." stands for transcriptions.}
\vspace{-10pt}
\scalebox{0.9}{
    \setlength{\tabcolsep}{4pt} % column spacing
    \renewcommand{\arraystretch}{1.3} % line hight
    \begin{tabular}{l|c|ccccc}
    \bottomrule
    Trans. & WER & Accuracy & Precision & Recall & F1 & AUC  \\
    \noalign{\hrule height 0.9pt}
    Manual & $-$ & \textbf{0.88} & $0.91$ & $0.83$ & $0.87$ & $0.88$ \\
    \hline
    ASR & $33.9\%$ & \textbf{0.88} & $0.82$ & $0.96$ & $0.88$ & $0.88$ \\
    \toprule
    \end{tabular}
}
% \caption{AD detection performance using Manual and ASR transcription.
% ``Trans." stands for transcriptions.}
\label{tab:basic_acc}
\end{table}

% \hfill
\begin{table}[htbp]
% \vspace{-5pt}
\centering
\caption{Confusion matrix of AD detection results.
``TP, TN, FP, FN" stands for true positive, true negative, false positive and false negative, respectively.}
\vspace{-10pt}
\scalebox{0.9}{
    \setlength{\tabcolsep}{10pt} % column spacing
    \renewcommand{\arraystretch}{1.3} % line hight
    \begin{tabular}{l|ccccc}
    \bottomrule
    Transcriptions & TP & TN & FP & FN & Total   \\
    \noalign{\hrule height 0.9pt}
    Manual & 20 & 22 & 2 & 4 & 48 \\
    \hline
    ASR & 23 & 19 & 5 & 1 & 48\\
    \toprule
    \end{tabular}
}
\vspace{-15pt}
\label{tab:basic_confusion}
\end{table}

\vspace{-0.1cm}
Table~\ref{tab:basic_confusion} further breaks down prediction results into confusion matrices.
It reflects that although two types of transcriptions achieve the same accuracy, prediction with ASR transcriptions tends \textit{more false positive errors}.
Compared to manual transcription, using ASR transcriptions results in 7 additional positive predictions (3 cases changed from true negatives to false positives, 4 cases changed from false negatives to true positives), and only one more negative prediction (a case changed from false negative to true positive).
This suggests that transcriptions with ASR errors might be more likely to be classified as originating from AD patients.
However, there are 40/48 ASR-robust cases where manual and ASR transcriptions yield the same classification results, despite the high word error rate of 30\%.
This raises questions about how these errors influence downstream decision-making.

\vspace{-0.2cm}

% For a better understanding of this question, Table~\ref{tab:case} showcases two examples that manual and ASR transcriptions have consistent and inconsistent predictions.
% In both these two examples, ASR transcriptions have a large proportion of word errors in a description of a specific scenario of the Cookie Theft picture, but case s177 still holds the true negative prediction.
% As for these examples, one hypothesis is that the word errors in s207 largely exist on the content words of the described picture (i.e., \textit{stool}, \textit{tipping}, \textit{mother}, \textit{washing}), while most errors in s177 lay on many non-content words including \textit{at}, \textit{him}, \textit{of}, \textit{the}, etc.

% \input{tables/showcase}

\begin{figure}[htbp]
\centering
\makebox[1\columnwidth][c]{
\hspace{-0.5cm}
\includegraphics[width=1.05\columnwidth]{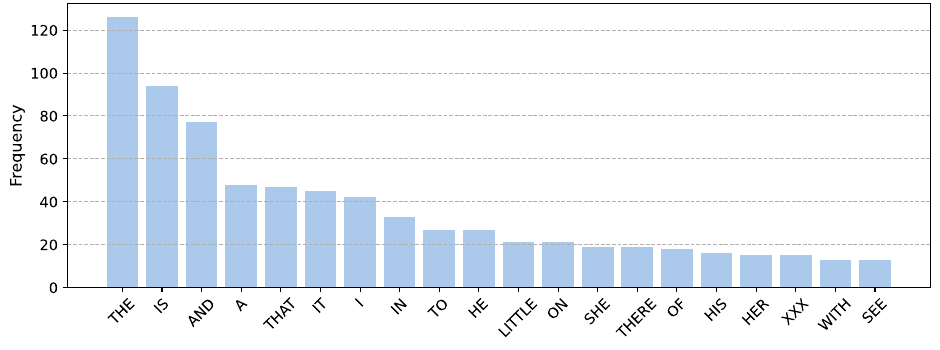}
}
\vspace*{-24pt}
\caption{
    The top 20 error word distribution of the ASR system.
}
\vspace{-12pt}
\label{fig:20-errors}
\end{figure}

\vspace{-0.2cm}

\section{Analysis of speech recognition errors}

Based on the detection system in Section~\ref{sec:base}, this section characterizes the composition of ASR errors and investigates why a high word error rate did not significantly degrade AD detection performance. 
We will start with error distribution and then examine certain types of errors using the test set transcriptions.

\vspace{-0.2cm}
\subsection{Basic analysis}
\vspace{-0.1cm}
\label{sec:basic}

The classification results show that there are 40 out of 48 cases (ASR-robust cases) where ASR and manual transcriptions yielded identical predictions.
To look into details, we first draw the distribution of the top 20 ASR errors.
Given that these errors did not lead to prediction changes, we assume that frequent errors play a relatively marginal role in AD detection.
As shown in Figure~\ref{fig:20-errors}, we can first observe these errors follow a long-tail distribution, with a small proportion of words constituting a large proportion of errors.
Moreover, we noticed that most of these top word errors are stopwords, which have been considered to carry less semantic weight and are not directly related to the picture description task used to diagnose AD patients.

\begin{figure}[tbp]
\centering
\makebox[1\columnwidth][c]{
\hspace{-0.5cm}
\includegraphics[width=1.05\columnwidth]{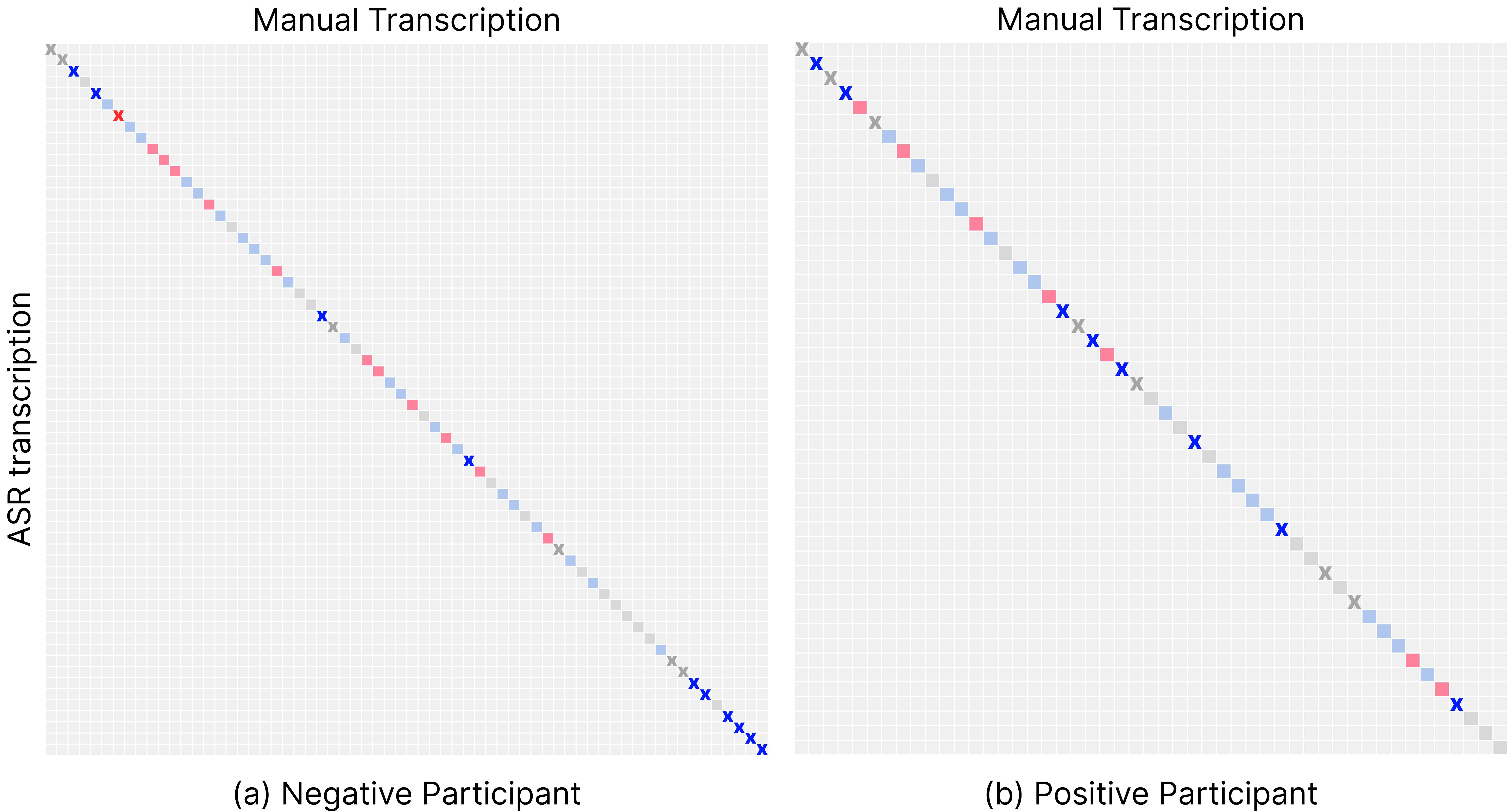}
}
\vspace*{-15pt}
\caption{
    Samples of alignment map for participants \textit{s170} (a) and \textit{s179} (b). 
    Squares represent correctly transcribed words, while crosses ('x') indicate ASR errors.
    Blue color for stopwords, red for keywords, and gray for other words. 
}
% \vspace{-0.7cm}
\label{fig:heatmap}
\end{figure}

In addition to overall errors, we conducted a thorough case study to observe the individual patterns of ASR errors.
Particularly, we generated alignment maps for each participant to visualize the compositions of ASR transcriptions and their discrepancy with manual transcription.
As showcased in Figure~\ref{fig:heatmap}, the alignment map compares manual and ASR transcriptions word-by-word, where correct words are marked as squares and errors are marked as crosses.
Following the findings in the previous section, we highlight the low-semantic stopwords\footnote{Following NLTK~\cite{bird-loper-2004-nltk} (v3.8.1) Stopwords List} with blue colors.
In comparison, we additionally highlight the relatively high-semantic task-related keywords with red colors.
As the diagnosis of AD in the ADReSS corpus is based on a picture description task, we hypothesize that content words related to the elements in the target picture may play a role in distinguishing AD patients from healthy controls.
Table~\ref{tab:kwlist} lists 39 task-related keywords used in this experiment, drawing on study~\cite{li2024useful}.

Going through the alignment maps of ASR-robust cases, we have the following findings.
First, stopwords and keywords covered most of the words in transcriptions, thus investigations of these words considered a majority of content.
Second, healthy participants generally could mention two times of keywords than participants with AD, which was reflected by more red markers in their alignment maps.
Additionally, aligning with the error distribution, a large proportion of ASR errors are on stopwords, which means \textit{most keywords are well preserved in ASR transcriptions}.
These observations suggest that ASR errors are dominated by low-semantic stopwords, while keywords related to AD diagnosis are relatively well recognized. 
This may explain why these errors have a limited impact on the final AD classification. 
The following part will present our detailed investigation of stopwords and keywords.

% WER	ACC	P	R	F	A
% Manual	0	0.88	0.85	0.92	0.88	0.87
% TDNN	33.9	0.88	0.82	0.96	0.88	0.88
% Whisper	35.4	0.85	0.77	1	0.87	0.85

% \vspace{-10pt}
\begin{table}[tbp]
\vspace{-5pt}
\centering
\caption{List of the task-related keywords}
\vspace{-10pt}
\scalebox{0.9}{
\setlength{\tabcolsep}{4pt} % column spacing
\renewcommand{\tabularxcolumn}[1]{>{\raggedright\arraybackslash}p{#1}}
\begin{tabular}{c}

\bottomrule

Task-related Keywords \\
\hline
\makecell[l]{boy, girl, two, children, cookie, jar, brother, sister. steal, cupboard, \\
door, finger, women, lady, dry, washing, dish, water, overflow,\\
floor, kitchen, window, curtain, mother, plate, stand, sink, running, \\
fall, outside, tree, house, look, open, hand, glass}\\

\toprule

\end{tabular}
}
\label{tab:kwlist}
\vspace{-15pt}
\end{table}

\vspace{-0.2cm}
\subsection{Anayisis of stopwords}
\vspace{-0.1cm}
\label{sw}
A series of experiments was conducted to examine the role of stopwords in AD detection.
First, Table \ref{tab:stopwords} presents a detailed breakdown of stopwords and non-stopwords in ASR errors from the ASR-robust cases.
It shows that a small proportion of stopword types (24\%) constitute the majority of errors (60\%), while the remaining 76\% types of words constitute only 40\% of errors. 
This aligns with the observation in Figure \ref{fig:20-errors} and \ref{fig:heatmap}.
Given these errors did not affect AD classification, we hypothesize that when using BERT features, the misrecognized stopwords error may not significantly propagate to the downstream detection decision.
Meanwhile, we note that the non-stopword errors also show an insignificant impact on classification results, suggesting that words other than stopwords could also play a marginal role in distinguishing AD.
We will discuss this further in the later part of this paper.

% \vspace{-10pt}
\begin{table}[tbp]
\hspace{-.35cm}
\centering
\caption{Word error rates (WER) and counts (ratio \%) of stopwords and non-stopwords in manual transcription (Counts) and ASR errors (Errors Counts). 
``tokens" represents the count of words and ``types" represents the count of distinct words.
}
\vspace{-5pt}
\scalebox{0.78}{
\setlength{\tabcolsep}{4pt} % column spacing
\renewcommand{\arraystretch}{1.2} % line hight
\newcolumntype{C}[1]{>{\centering\arraybackslash}p{#1}}
\begin{tabular}{cccccc}
\toprule
\multirow{2}{*}{} & \multirow{2}{*}{WER} & \multicolumn{2}{c}{Counts} & \multicolumn{2}{c}{Errors Counts} \\
\cmidrule(lr){3-4} \cmidrule(lr){5-6}
 & & tokens & types & tokens & types \\
\midrule
Stopwords & $32.4\%$ & $2574(58\%)$ &$95(18\%)$ & $834(\textbf{60\%})$ & $87(\textbf{24\%})$   \\
Non-stopwords& $29.8\%$ & $1897(42\%)$ & $447(82\%)$ & $566(\textbf{40\%})$ & $281(\textbf{76\%})$   \\

\bottomrule

\end{tabular}
}
\label{tab:stopwords}
\vspace{-15pt}
\end{table}

\begin{figure}[htbp]
\centering
\makebox[1\columnwidth][c]{
\hspace{-0.5cm}
\includegraphics[width=1.0\columnwidth]{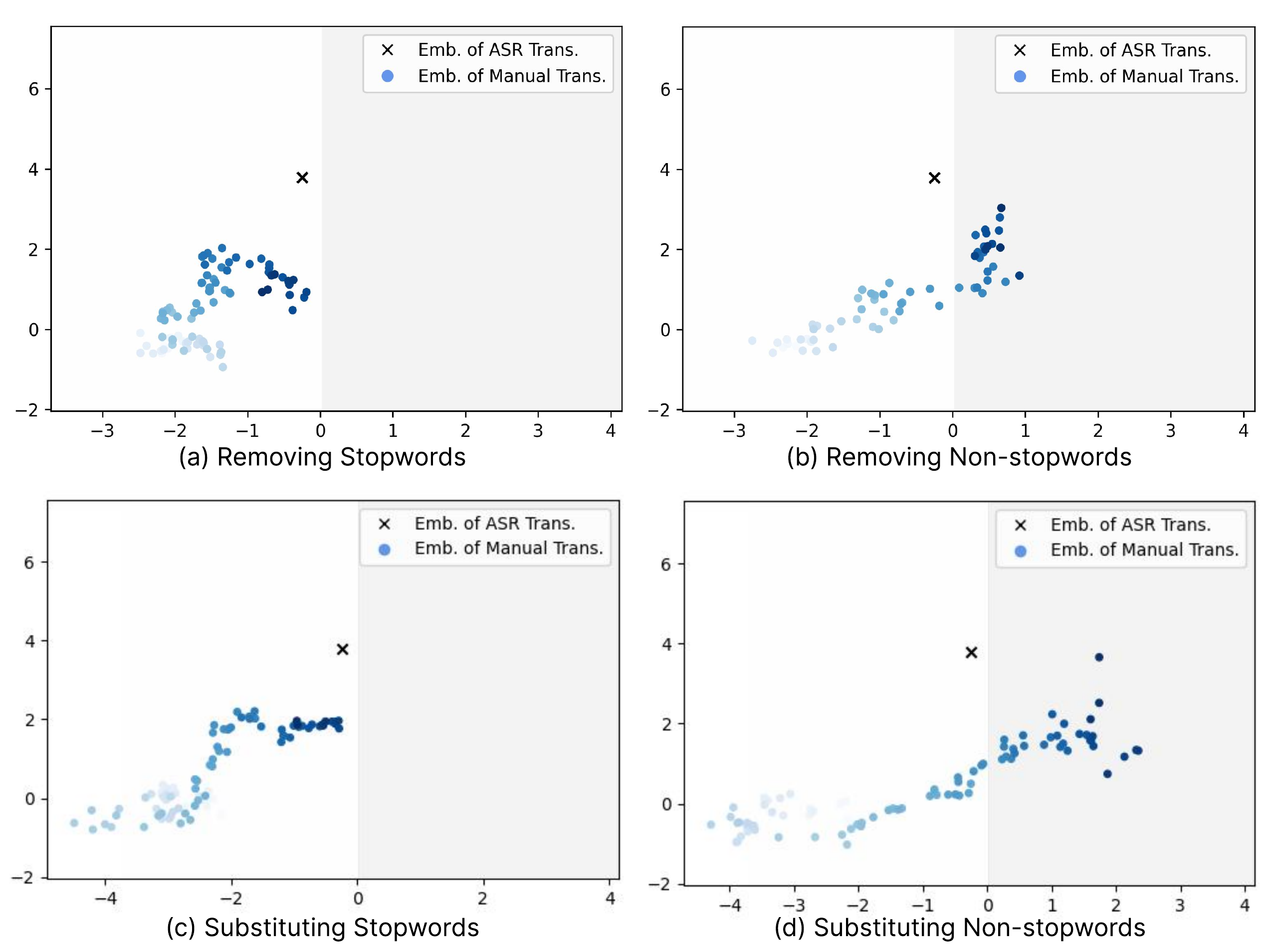}
}
\vspace*{-15pt}
\caption{
    BERT embedding variations as stopwords are incrementally removed/substituted from manual transcriptions (participant s172). 
    Light to dark blue: fewer to more stopwords removed/substituted. 
    'x': ASR transcript.
    White/gray: health/AD decision regions.
}
\vspace{-5pt}
\label{fig:ablation-sw}

\end{figure}

\noindent \textbf{Words ablation experiments}
To verify the impact of stopwords on downstream classification, we conduct a word ablation experiment to probe the variation of BERT embeddings when stopwords are misrecognized.
Starting from manual transcriptions, we randomly removed or substituted 10\% to 100\% of stopwords in 10\% increments and observed the corresponding BERT embeddings. 
For substitution, we replaced random stopwords with an equal number of non-stopwords randomly sampled from all manual transcriptions.
We then projected the high-dimensional BERT embeddings of the edited transcriptions onto a 2-dimensional subspace orthogonal to the original SVM decision hyperplane. 
The two sides of the decision boundary were filled with white (healthy) and gray (AD) colors to demonstrate the relationship between BERT embeddings and SVM decisions.
Figure~\ref{fig:ablation-sw} shows the resulting visualizations from an example participant.
When gradually removing or substituting stopwords (Figure~\ref{fig:ablation-sw} (a) \& (c)), BERT embeddings show a shift towards the decision boundary.
This suggests that stopwords may have an effect on AD detection. 
However, the edited transcription still falls on the negative (recognized as healthy) side until all stopwords are removed/substituted. 
In contrast, when gradually removing/substituting the same number of non-stopwords (Figure~\ref{fig:ablation-sw} (b) \& (d)), the embeddings shift across the decision boundary with a more direct trend.
Moreover, in this case, substitutions result in a larger embedding variance compared to removal. 
We attribute this to the randomness in the substitution operation, as random words were sampled for replacement.

\vspace{-10pt}

\begin{figure}[tbp]
\centering
\makebox[1\columnwidth][c]{
\hspace{-0.5cm}
\includegraphics[width=1.0\columnwidth]{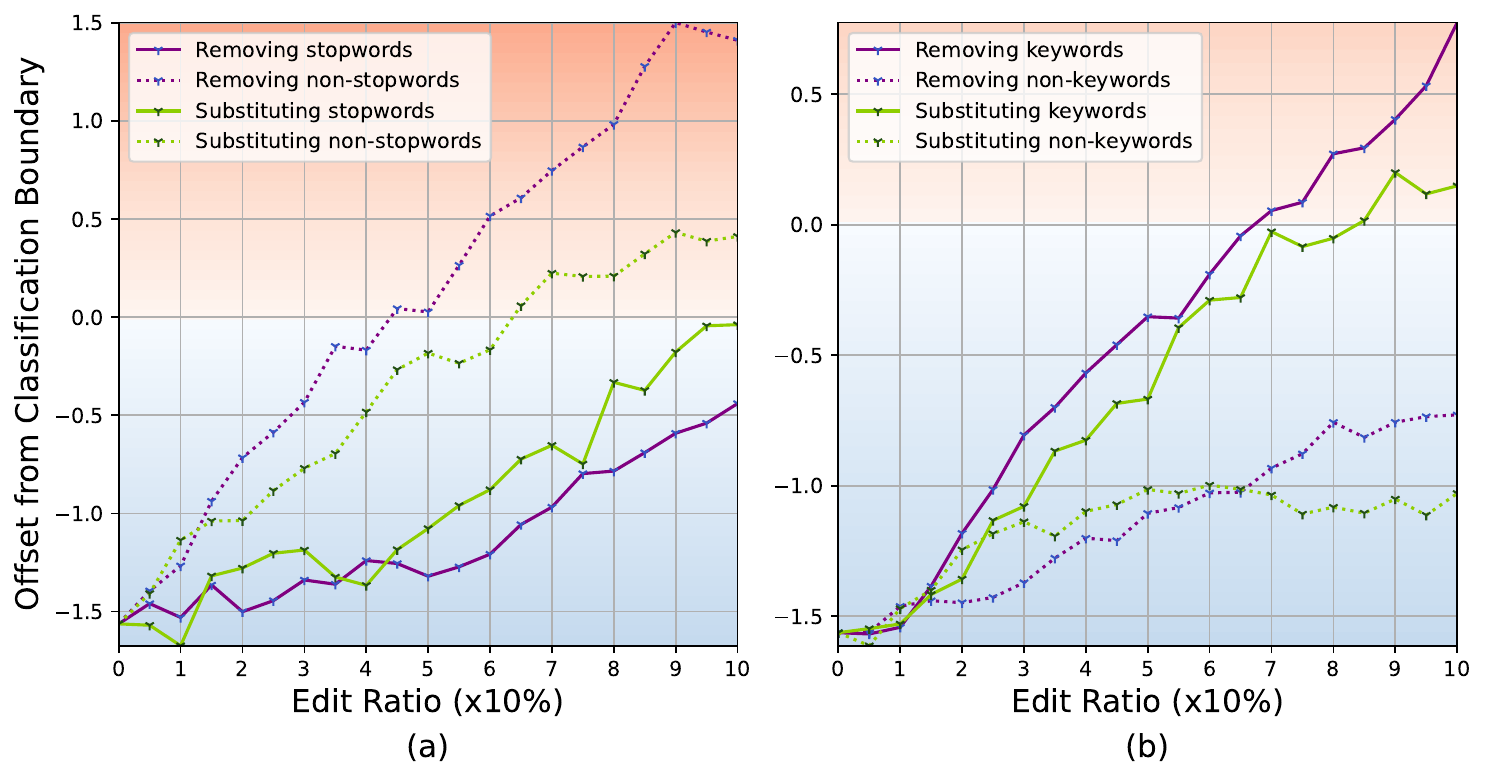}
}
\vspace*{-20pt}
\caption{
    Average hyperplane offset for transcription embeddings as stopwords (a) and keywords (b) are incrementally removed or substituted. Edit ratio represents the percentage of words removed/substituted.
}
\vspace{-18pt}
\label{fig:curves}
\end{figure}

Furthermore, to reduce randomness and quantify this experiment across participants, we calculate the \textit{hyperplane offset} for each participant and each removal/substitution.
This offset represents a signed distance from BERT embedding to the SVM hyperplane.
\begin{equation} \label{eq}
\begin{aligned}
d=(w^Tx+b)/||w||
\end{aligned}
\end{equation}
The above equation presents calculation details where $w$ and $b$ are the norm vector and interception of SVM, $x$ is the input vector, and $||\cdot||$ stands for L2 norm.
The resulting $d$ reflects how embeddings move toward or against the decision boundary.
We averaged $d$ across participants to observe the average embedding movement when transcriptions are edited.
In Figure~\ref{fig:curves} (a), we present the resulting offset curves, using blue and red background colors to indicate healthy and AD decision regions, respectively.
Similar to the embedding visualization, when removing or substituting stopwords (solid lines), the average offset values gradually approach 0 from negative values but do not cross the decision boundary.
Suggesting that the edit of stopwords leads to embeddings shift towards the AD decision region, but not enough to change the classification outcome.
In contrast, editing non-stopwords easily leads to set values reaching 0, indicating a likely change in the classification result.
This difference implies distinct roles of stopwords and non-stopwords in the representations, with non-stopwords carrying more semantic content that influences the AD detection task.

\vspace{-0.2cm}
\subsection{Analysis of task-related keywords}
\vspace{-0.1cm}
In comparison to stopwords, this part investigates keywords related to elements in the Cookie Thief picture used for AD diagnosis.
Table \ref{tab:keywords} breakdown keyword and non-keyword errors of ASR-robust cases.
We highlight that the WER for keywords is only 14.3\%, which is much lower than the overall WER of 33.9\%.
The high accuracy in keywords is understandable, as task-related keywords are frequently mentioned words in the training set, with only 60\footnote{Including inflections of words in Table \ref{tab:kwlist}} distinct keywords in total.
Compared to stopwords, most keywords have more syllables, making them easier to recognize.
Furthermore, keyword errors account for just 9\% of all ASR errors.
This aligns with the observations using the alignment maps in Section \ref{sec:basic}.

\begin{figure}[htbp]
\centering
\makebox[1\columnwidth][c]{
\hspace{-0.5cm}
\includegraphics[width=1.0\columnwidth]{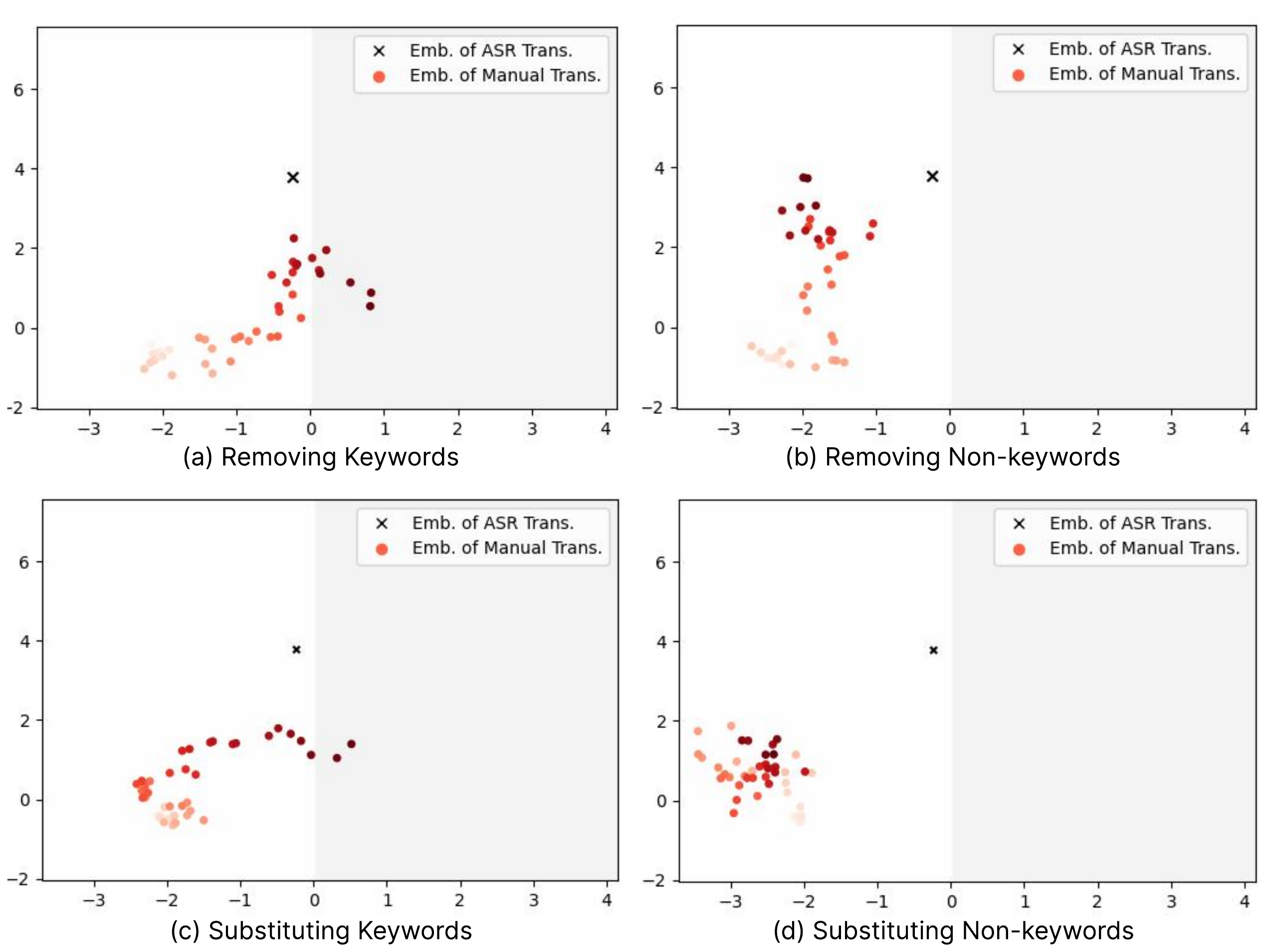}
}
\vspace*{-0.55cm}
\caption{
    BERT embedding variations as keywords are incrementally removed/substituted from manual transcriptions (participant s172). 
    Light to dark blue: fewer to more keywords removed/substituted. 
    'x': ASR transcript.
    White/gray: health/AD decision regions.
}
\vspace{-0.2cm}
\label{fig:ablation-kw}
\end{figure}

\noindent \textbf{Words ablation experiments}
We conducted word ablation experiments using the same settings as in the stopword investigation, replacing stopwords with keywords.
The visualizations of BERT embeddings are exampled in Figure \ref{fig:ablation-kw}.
It is not surprising that the results of editing keywords and stopwords are opposite.
Removing or substituting keywords in transcriptions causes BERT embeddings to shift across the SVM decision boundary, whereas similar manipulations of non-keywords do not produce this effect.
Figure \ref{fig:curves} (b) depicted the hyperplane offset experiment.
There is also a clear divergence when editing keywords and non-keywords.
Opposite to stopwords, removing or substituting keywords shows more significant changes in embedding offset.
This divergence indicates \textit{keywords may play a distinctly meaningful role compared to other words}.
Some of the 88\% non-keywords may also be non-stopwords, which could explain the insignificant non-stopwords in Table \ref{tab:stopwords}.
Notably, editing keywords didn't result in embedding shifts as effectively as expected, i.e., the averaged offset started to be positive only when 70\% of the keywords were removed. 
We interpret this as an indication that there exists information above the word level that can be helpful to AD classification.
E.g., sentence length and the interplay between words within the transcriptions.

% \vspace{-0.15cm}

% \vspace{-10pt}
\begin{table}[tbp]
\hspace{-.35cm}
\centering
\caption{Word error rates (WER) and counts (ratio \%) of keywords and non-keywords in manual transcription (Counts) and ASR errors (Errors Counts). 
``tokens" represents the count of words and ``types" represents the count of distinct words.
}
\vspace{-8pt}
\scalebox{0.78}{
\setlength{\tabcolsep}{4pt} % column spacing
\renewcommand{\arraystretch}{1.2} % line hight
% \toprule
% \end{tabular}
\newcolumntype{C}[1]{>{\centering\arraybackslash}p{#1}}
\begin{tabular}{cccccc}
\toprule
\multirow{2}{*}{} & \multirow{2}{*}{WER} & \multicolumn{2}{c}{Counts} & \multicolumn{2}{c}{Errors Counts} \\
\cmidrule(lr){3-4} \cmidrule(lr){5-6}
 & & tokens & types & tokens & types \\
\midrule
Keywords & $\textbf{14.3\%}$ & $909(20\%)$ & $60(11\%)$ & $130(\textbf{9\%})$ & $46(12\%)$   \\
Non-keywords & $\textbf{35.6\%}$ & $3562(80\%)$ & $482(89\%)$ & $1270(\textbf{91\%})$ & $322(88\%)$   \\

\bottomrule

\end{tabular}
}
\label{tab:keywords}
\vspace{-15pt}
\end{table}
\vspace{-10pt}

\vspace{-0.3cm}

\vspace{-0.15cm}
\section{Conclusions}

We performed a series of analyses to investigate the effect of ASR transcription errors in BERT-based Alzheimer's Disease (AD) detection. 
Specifically, we explored why transcriptions with notable word error rates could yield detection accuracy equivalent to that of manual transcriptions.
We have shown that not all ASR errors are equally detrimental to downstream AD detection.
Certain word types, particularly stopwords, constitute a large proportion of errors while carrying limited information for AD discrimination than others.
The task-related keywords were demonstrated to be significant to AD classification while occupying only 9\% of all word errors.
These findings provide insights into understanding the influence of ASR errors on downstream AD detection and facilitate the development of robust AD detection systems.
While this work focused on ASR at the word level, future work could explore their influence on other linguistic structures such as syntactic dependencies and discourse coherence.

\vspace{-0.15cm}
\section{Acknowledgements} \label{sec:ack}
This work is supported by the HKSARG Research Grants Council’s Theme-based Research Grant Scheme (Project No. T45- 407/19N) and the CUHK Stanley Ho Big Data Decision Research Centre.

\bibliographystyle{IEEEtran}
\bibliography{mybib}

\begin{thebibliography}{10}
\providecommand{\url}[1]{#1}
\csname url@samestyle\endcsname
\providecommand{\newblock}{\relax}
\providecommand{\bibinfo}[2]{#2}
\providecommand{\BIBentrySTDinterwordspacing}{\spaceskip=0pt\relax}
\providecommand{\BIBentryALTinterwordstretchfactor}{4}
\providecommand{\BIBentryALTinterwordspacing}{\spaceskip=\fontdimen2\font plus
\BIBentryALTinterwordstretchfactor\fontdimen3\font minus \fontdimen4\font\relax}
\providecommand{\BIBforeignlanguage}[2]{{%
\expandafter\ifx\csname l@#1\endcsname\relax
\typeout{** WARNING: IEEEtran.bst: No hyphenation pattern has been}%
\typeout{** loaded for the language `#1'. Using the pattern for}%
\typeout{** the default language instead.}%
\else
\language=\csname l@#1\endcsname
\fi
#2}}
\providecommand{\BIBdecl}{\relax}
\BIBdecl

\bibitem{lynch2020world}
C.~Lynch, ``World alzheimer report 2019: Attitudes to dementia, a global survey: Public health: Engaging people in adrd research,'' \emph{Alzheimer's \& Dementia}, 2020.

\bibitem{nasreddine2005montreal}
Z.~S. Nasreddine, N.~A. Phillips, V.~B{\'e}dirian, S.~Charbonneau, V.~Whitehead, I.~Collin, J.~L. Cummings, and H.~Chertkow, ``The montreal cognitive assessment, {MoCA}: a brief screening tool for mild cognitive impairment,'' \emph{Journal of the American Geriatrics Society}, vol.~53, no.~4, pp. 695--699, 2005.

\bibitem{reilly2004frog}
J.~Reilly, M.~Losh, U.~Bellugi, and B.~Wulfeck, ````frog, where are you?'' narratives in children with specific language impairment, early focal brain injury, and williams syndrome,'' \emph{Brain and language}, vol.~88, no.~2, pp. 229--247, 2004.

\bibitem{martinez2021ten}
I.~Mart{\'\i}nez-Nicol{\'a}s, T.~E. Llorente, F.~Mart{\'\i}nez-S{\'a}nchez, and J.~J.~G. Meil{\'a}n, ``Ten years of research on automatic voice and speech analysis of people with alzheimer's disease and mild cognitive impairment: a systematic review article,'' \emph{Frontiers in Psychology}, vol.~12, p. 620251, 2021.

\bibitem{meng2023integrated}
H.~Meng, B.~Mak, M.-W. Mak, H.~Fung, X.~Gong, T.~Kwok, X.~Liu, V.~Mok, P.~Wong, J.~Woo \emph{et~al.}, ``Integrated and enhanced pipeline system to support spoken language analytics for screening neurocognitive disorders,'' \emph{Interspeech}, 2023.

\bibitem{alhanai2017spoken}
T.~Alhanai, R.~Au, and J.~Glass, ``Spoken language biomarkers for detecting cognitive impairment,'' 2017.

\bibitem{fraser2016linguistic}
K.~C. Fraser, J.~A. Meltzer, and F.~Rudzicz, ``Linguistic features identify alzheimer’s disease in narrative speech,'' \emph{Journal of Alzheimer's Disease}, 2016.

\bibitem{weiner2019speech}
J.~Weiner, C.~Frankenberg, J.~Schr{\"o}der, and T.~Schultz, ``Speech reveals future risk of developing dementia: Predictive dementia screening from biographic interviews,'' in \emph{ASRU}.\hskip 1em plus 0.5em minus 0.4em\relax IEEE, 2019.

\bibitem{frankenberg2021verbal}
C.~Frankenberg, J.~Weiner, M.~Knebel, A.~Abulimiti, P.~Toro, C.~J. Herold, T.~Schultz, and J.~Schr{\"o}der, ``Verbal fluency in normal aging and cognitive decline: Results of a longitudinal study,'' \emph{Computer Speech \& Language}, 2021.

\bibitem{koo2020exploiting}
J.~Koo, J.~H. Lee, J.~Pyo, Y.~Jo, and K.~Lee, ``Exploiting multi-modal features from pre-trained networks for alzheimer's dementia recognition,'' in \emph{INTERSPEECH}, 2020.

\bibitem{balagopalan2021comparingPretrained}
A.~Balagopalan, B.~Eyre, J.~Robin, F.~Rudzicz, and J.~Novikova, ``Comparing pre-trained and feature-based models for prediction of alzheimer's disease based on speech,'' \emph{Frontiers in aging neuroscience}, 2021.

\bibitem{syed2021automated}
Z.~S. Syed, M.~S.~S. Syed, M.~Lech, and E.~Pirogova, ``Automated recognition of alzheimer’s dementia using bag-of-deep-features and model ensembling,'' \emph{IEEE Access}, 2021.

\bibitem{balagopalan2020bert}
A.~Balagopalan, B.~Eyre, F.~Rudzicz, and J.~Novikova, ``To bert or not to bert: comparing speech and language-based approaches for alzheimer's disease detection,'' \emph{arXiv preprint arXiv:2008.01551}, 2020.

\bibitem{yuan2020disfluencies}
J.~Yuan, Y.~Bian, X.~Cai, J.~Huang, Z.~Ye, and K.~Church, ``Disfluencies and fine-tuning pre-trained language models for detection of alzheimer's disease.'' in \emph{INTERSPEECH}, 2020.

\bibitem{martinc2021temporal}
M.~Martinc, F.~Haider, S.~Pollak, and S.~Luz, ``Temporal integration of text transcripts and acoustic features for alzheimer's diagnosis based on spontaneous speech,'' \emph{Frontiers in Aging Neuroscience}, 2021.

\bibitem{devlin2019bert}
J.~Devlin, M.-W. Chang, K.~Lee, and K.~Toutanova, ``{BERT}: Pre-training of deep bidirectional transformers for language understanding,'' 2019.

\bibitem{liu2019roberta}
Y.~Liu, M.~Ott, N.~Goyal, J.~Du, M.~Joshi, D.~Chen, O.~Levy, M.~Lewis, L.~Zettlemoyer, and V.~Stoyanov, ``Roberta: A robustly optimized bert pretraining approach,'' 2019.

\bibitem{syed2020automated}
M.~S.~S. Syed, Z.~S. Syed, M.~Lech, and E.~Pirogova, ``Automated screening for alzheimer's dementia through spontaneous speech.'' in \emph{Interspeech}, vol. 2020, 2020, pp. 2222--6.

\bibitem{li2021comparative}
J.~Li, J.~Yu, Z.~Ye, S.~Wong, M.~Mak, B.~Mak, X.~Liu, and H.~Meng, ``A comparative study of acoustic and linguistic features classification for alzheimer's disease detection,'' in \emph{ICASSP}.\hskip 1em plus 0.5em minus 0.4em\relax IEEE, 2021, pp. 6423--6427.

\bibitem{li2023leveraging}
J.~Li, K.~Song, J.~Li, B.~Zheng, D.~Li, X.~Wu, X.~Liu, and H.~Meng, ``Leveraging pretrained representations with task-related keywords for alzheimer's disease detection,'' 2023.

\bibitem{wang2022exploring}
Y.~Wang, T.~Wang, Z.~Ye, L.~Meng, S.~Hu, X.~Wu, X.~Liu, and H.~Meng, ``Exploring linguistic feature and model combination for speech recognition based automatic ad detection,'' \emph{INTERSPEECH}, 2022.

\bibitem{qiao2021alzheimer}
Y.~Qiao, X.~Yin, D.~Wiechmann, and E.~Kerz, ``Alzheimer’s disease detection from spontaneous speech through combining linguistic complexity and (dis) fluency features with pretrained language models,'' in \emph{INTERSPEECH}, 2021.

\bibitem{zhou2016speech}
L.~Zhou, K.~C. Fraser, and F.~Rudzicz, ``Speech recognition in alzheimer's disease and in its assessment.'' in \emph{Interspeech}, vol. 2016, 2016, pp. 1948--1952.

\bibitem{mirheidari2019dementia}
B.~Mirheidari, D.~Blackburn, T.~Walker, M.~Reuber, and H.~Christensen, ``Dementia detection using automatic analysis of conversations,'' \emph{Computer Speech \& Language}, vol.~53, pp. 65--79, 2019.

\bibitem{ye2021development}
Z.~Ye, S.~Hu, J.~Li, X.~Xie, M.~Geng, J.~Yu, J.~Xu, B.~Xue, S.~Liu, X.~Liu \emph{et~al.}, ``Development of the cuhk elderly speech recognition system for neurocognitive disorder detection using the dementiabank corpus,'' in \emph{ICASSP}.\hskip 1em plus 0.5em minus 0.4em\relax IEEE, 2021, pp. 6433--6437.

\bibitem{geng2022investigation}
M.~Geng, X.~Xie, S.~Liu, J.~Yu, S.~Hu, X.~Liu, and H.~Meng, ``Investigation of data augmentation techniques for disordered speech recognition,'' \emph{arXiv preprint arXiv:2201.05562}, 2022.

\bibitem{jin2023personalized}
Z.~Jin, M.~Geng, J.~Deng, T.~Wang, S.~Hu, G.~Li, and X.~Liu, ``Personalized adversarial data augmentation for dysarthric and elderly speech recognition,'' \emph{IEEE/ACM Transactions on Audio, Speech, and Language Processing}, 2023.

\bibitem{deng2021bayesian}
J.~Deng, F.~R. Gutierrez, S.~Hu, M.~Geng, X.~Xie, Z.~Ye, S.~Liu, J.~Yu, X.~Liu, and H.~Meng, ``Bayesian parametric and architectural domain adaptation of lf-mmi trained tdnns for elderly and dysarthric speech recognition.'' in \emph{Interspeech}, 2021, pp. 4818--4822.

\bibitem{wang2022conformer}
T.~Wang, J.~Deng, M.~Geng, Z.~Ye, S.~Hu, Y.~Wang, M.~Cui, Z.~Jin, X.~Liu, and H.~Meng, ``Conformer based elderly speech recognition system for alzheimer's disease detection,'' \emph{arXiv preprint arXiv:2206.13232}, 2022.

\bibitem{li2022far}
C.~Li, T.~Cohen, and S.~Pakhomov, ``The far side of failure: Investigating the impact of speech recognition errors on subsequent dementia classification,'' \emph{arXiv preprint arXiv:2211.07430}, 2022.

\bibitem{li2024useful}
C.~Li, W.~Xu, T.~Cohen, and S.~Pakhomov, ``Useful blunders: Can automated speech recognition errors improve downstream dementia classification?'' \emph{Journal of Biomedical Informatics}, vol. 150, p. 104598, 2024.

\bibitem{balagopalan2020impact}
A.~Balagopalan, K.~Shkaruta, and J.~Novikova, ``Impact of asr on alzheimer’s disease detection: All errors are equal, but deletions are more equal than others,'' in \emph{W-NUT}, 2020, pp. 159--164.

\bibitem{luz2020alzheimer}
S.~Luz, F.~Haider, S.~de~la Fuente, D.~Fromm, and B.~MacWhinney, ``{Alzheimer’s Dementia Recognition Through Spontaneous Speech: The ADReSS Challenge},'' \emph{INTERSPEECH}, 2020.

\bibitem{becker1994natural}
J.~T. Becker, F.~Boiler, O.~L. Lopez, J.~Saxton, and K.~L. McGonigle, ``The natural history of alzheimer's disease: description of study cohort and accuracy of diagnosis,'' \emph{Archives of neurology}, 1994.

\bibitem{kaplan1983boston}
E.~Kaplan, H.~Goodglass, and S.~Weintraub, \emph{Boston Naming Test}.\hskip 1em plus 0.5em minus 0.4em\relax Philadelphia, PA: Lea \& Febiger, 1983.

\bibitem{bird-loper-2004-nltk}
\BIBentryALTinterwordspacing
S.~Bird and E.~Loper, ``{NLTK}: The natural language toolkit,'' in \emph{Proceedings of the {ACL} Interactive Poster and Demonstration Sessions}.\hskip 1em plus 0.5em minus 0.4em\relax Barcelona, Spain: Association for Computational Linguistics, Jul. 2004, pp. 214--217. [Online]. Available: \url{https://aclanthology.org/P04-3031}
\BIBentrySTDinterwordspacing

\end{thebibliography}

\end{document}